\documentclass[a4paper,fleqn]{cas-dc}
\usepackage{soul}
\usepackage{cite}
\usepackage{float}
\usepackage{graphicx}
\usepackage{amsmath,amssymb,amsfonts}
\usepackage{algorithmic}
\usepackage{graphicx}
\usepackage{algorithm,algorithmic}
\usepackage{hyperref}
\hypersetup{hidelinks=true}
\usepackage{textcomp}
\usepackage{booktabs}
\usepackage{multicol}
\usepackage{makecell}
\usepackage{multirow}
\usepackage[authoryear,longnamesfirst]{natbib}
\usepackage{tabularray}
\usepackage{tabularx}
\def\tsc#1{\csdef{#1}{\textsc{\lowercase{#1}}\xspace}}
\tsc{WGM}
\tsc{QE}
\tsc{EP}
\tsc{PMS}
\tsc{BEC}
\tsc{DE}

\begin{document}
\let\WriteBookmarks\relax
\def\floatpagepagefraction{1}
\def\textpagefraction{.001}

\shorttitle{RetinaRegNet: A Zero-Shot Approach for Retinal Image Registration}

\shortauthors{Sivaraman et~al.}

\title [mode = title]{RetinaRegNet: A Zero-Shot Approach for Retinal Image Registration}                      


%
\author[1]{Vishal Balaji Sivaraman}
\credit{Methodology, Software, Formal analysis, Investigation, Data curation, Writing - original draft, Visualization}
\author[2]{Muhammad Imran}
\credit{Methodology, Software, Writing - review \& editing, Visualization}
\author[3]{Qingyue Wei}
\credit{Methodology, Writing - review \& editing, Visualization}
\author[4]{Preethika Muralidharan}
\credit{Data curation, Writing - review \& editing}
\author[5]{Michelle R. Tamplin}
\credit{Data curation, Writing - review \& editing}
\author[5]{Isabella M. Grumbach}
\credit{Data curation, Writing - review \& editing}
\author[5]{Randy H. Kardon}
\credit{Data curation, Supervision, Writing - review \& editing}
\author[6]{Jui-Kai Wang}
\credit{Data curation, Supervision, Writing - review \& editing}
\author[7]{Yuyin Zhou}
\credit{Methodology, Supervision, Writing review \& editing}
\author[1,2,4,8]{Wei Shao}[
orcid = 0000-0003-4931-4839]
\cormark[1]
\ead{weishao@ufl.edu}

\credit{Conceptualization, Methodology, Resources, Data curation, Writing - original draft, Supervision, Project administration, Funding acquisition}

\affiliation[1]{organization={Department of Electrical and Computer Engineering, University of Florida},
    city={Gainesville},
    state={Florida},
    postcode={32610}, 
    country={United States.}
    }

    \affiliation[2]{organization={Department of Medicine, University of Florida},
    city={Gainesville},
    state={Florida},
    postcode={32610}, 
    country={United States.}
    }
    
    \affiliation[3]{organization={Department of Computational and Mathematical Engineering, Stanford University},
    city={Stanford},
    state={California},
    postcode={94305}, 
    country={United States.}
    }
    
     \affiliation[4]{organization={Department of Health Outcomes and Biomedical Informatics, University of Florida},
    city={Gainesville},
    state={Florida},
    postcode={32610}, 
    country={United States.}
    }
     \affiliation[5]{organization={Iowa City VA Center for the Prevention and Treatment of Visual Loss, Department of Internal Medicine, University of Iowa},
    city={Iowa City},
    state={Iowa},
    postcode={52242}, 
    country={United States.}
    }
    \affiliation[6]{organization={Iowa City VA Center for the Prevention and Treatment of Visual Loss, Department of Ophthalmology and Visual Sciences, Department of Electrical and Computer Engineering, University of Iowa},
    city={Iowa City},
    state={Iowa},
    postcode={52242}, 
    country={United States.}
    }
    \affiliation[7]{organization={Department of Computer Science and Engineering, University of California, Santa Cruz},
        city={Santa Cruz},
        state={California},
        postcode={95064}, 
        country={United States.}
        }
    \affiliation[8]{organization={Intelligent Clinical Care Center, University of Florida},
        city={Gainesville},
        state={Florida},
        postcode={32610}, 
        country={United States.}
        }
        
\cortext[cor1]{Corresponding author}

\begin{abstract}
We introduce RetinaRegNet, a zero-shot image registration model designed to register retinal images with minimal overlap, large deformations, and varying image quality. RetinaRegNet addresses these challenges and achieves robust and accurate registration through the following steps. First, we extract features from the moving and fixed images using latent diffusion models. We then sample feature points from the fixed image using a combination of the SIFT algorithm and random point sampling. For each sampled point, we identify its corresponding point in the moving image using a 2D correlation map, which computes the cosine similarity between the diffusion feature vectors of the point in the fixed image and all pixels in the moving image. Second, we eliminate most incorrectly detected point correspondences (outliers) by enforcing an inverse consistency constraint, ensuring that correspondences are consistent in both forward and backward directions. We further remove outliers with large distances between corresponding points using a global transformation-based outlier detector. Finally, we implement a two-stage registration framework to handle large deformations. The first stage estimates a homography transformation to achieve global alignment between the images, while the second stage uses a third-order polynomial transformation to estimate local deformations. We evaluated RetinaRegNet on three retinal image registration datasets: color fundus images, fluorescein angiography images, and laser speckle flowgraphy images. Our model consistently outperformed state-of-the-art methods across all datasets. The accurate registration achieved by RetinaRegNet enables the tracking of eye disease progression, enhances surgical planning, and facilitates the evaluation of treatment efficacy. Our code is publicly available at: \url{https://github.com/mirthAI/RetinaRegNet}.

\end{abstract}

\begin{keywords}
Retinal image registration \sep Zero-shot image registration\sep Inverse consistency \sep Outlier detector
\end{keywords}

\maketitle

\section{Introduction}
\label{sec:introduction}

Image registration is a fundamental technique in medical imaging, with diverse applications including image-guided radiation therapy~\citep{foskey2005large}, disease progression monitoring~\citep{gorbunova2008weight}, motion tracking~\citep{shao2021geodesic}, brain mapping~\citep{toga2001role}, and image fusion~\citep{shao2021prosregnet}. The goal of image registration is to align a moving (source) image with a fixed (target) image by estimating a geometric transformation that matches corresponding features or structures in the two images. In this study, we focus on retinal image registration, which enables the monitoring of eye disease progression and the assessment of treatment effectiveness~\citep{saha2019color}. The main challenges in retinal image registration include estimating large deformations, registering images with minimal overlap, and ensuring robustness to variations in image quality~\citep{Can2002}. Large deformations involve the alignment of images with notable differences in position or scale, which often result in overfitting in transformation estimation or complicate point correspondence estimation. Small overlap between the fixed and moving images presents challenges since it can complicate the identification of corresponding features necessary for accurate alignment. Additionally, variations in image quality, caused by pathologies or differences in illumination, contrast, or noise levels, can further complicate the registration process, making it difficult to achieve a robust and accurate transformation. Overcoming these challenges requires the development of advanced algorithms that can handle complex transformations and remain robust to the inherent variability in retinal images.

To address the challenges in retinal image registration, we advocate for feature-based registration methods, starting with a feature point detector, followed by image feature extraction and matching, and transformation estimation. The most widely used feature point detector is the Scale-Invariant Feature Transform (SIFT) algorithm~\citep{lowe2004distinctive}, which can detect and describe rotation- and scaling-invariant local image features like corners and blobs. However, SIFT has limitations when applied to retinal image registration. First, most detected feature points tend to cluster in areas with rich textural features, leading to sparse or nonexistent point correspondences in homogeneous regions without distinctive features. Second, the image features generated by SIFT provide only a localized perspective, often leading to outliers in feature matching for image pairs with large deformations or minimal overlap.
Two primary approaches have been developed to overcome the limitations of SIFT. One approach focuses on developing more efficient feature detectors, including general-purpose detectors~\citep{bay2006surf, alcantarilla2012kaze} as well as those tailored for specific registration applications~\citep{wang2011bfsift, liu2022semi, nasser2023reverse}. The other approach is detector-free, involving the extraction of two-dimensional (2D) feature maps from both fixed and moving images to compute a four-dimensional (4D) correlation map~\citep{rocco2018neighbourhood}. This map facilitates the identification of more feature point correspondences in regions with fewer distinct features.

In this paper, we introduce RetinaRegNet, a zero-shot model that achieves state-of-the-art performance on various retinal image registration tasks through several innovations. First, we sampled feature points from regions with rich features (e.g., edges, corners) and regions without distinct features to ensure that feature points are distributed throughout the images. This significantly improved registration accuracy in homogeneous regions. Second, our model reliably identified point correspondences using the concept of the inverse consistency constraint~\citep{Pascal1992,christensen2001consistent}, which was originally used to ensure that if two images are registered to each other and then reversely registered, the estimated transformations are inverses of each other. In the context of estimating point correspondences between two images, inverse consistency is a necessary condition to ensure the correspondences are consistent in both the forward and backward directions. Third, our model excluded incorrect point correspondences where points have a long distance using a transformation-based outlier detector. The most widely used RANSAC (Random Sample Consensus) outlier-detection algorithm~\citep{fischler1981random} relies on random sampling, which can lead to inconsistent results and increased computational time, especially when dealing with complex data with many outliers. Finally, our model utilized a two-stage registration framework. The first stage involved estimating a homography deformation to achieve a global alignment of the two images. This enabled step two, which was the estimation of a more accurate non-rigid local transformation. This two-stage design enhanced both the robustness and accuracy of our model.

\begin{figure}[!htbp]
  \centering
\includegraphics[width=1.0\linewidth]{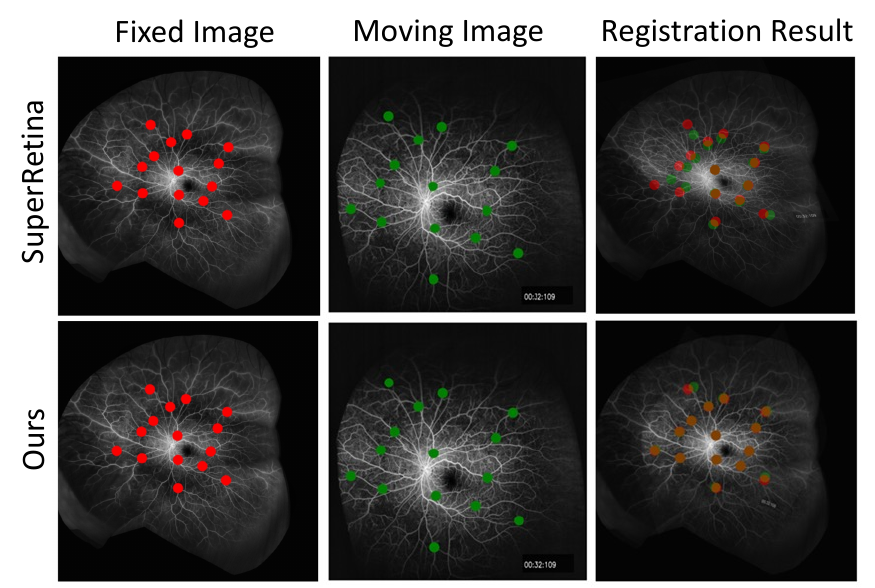}
  \caption{Visual comparison of the performance of RetinaRegNet and SuperRetina~\citep{liu2022semi} when registering an image pair with large displacement and scaling deformations. Red points represent landmarks on the fixed image, and green points represent the corresponding landmarks on the moving image. In the third column, the deformed images are overlaid with the fixed image, and landmarks on both the fixed and deformed images are shown in this overlay. The results demonstrate that our registration model significantly improves the alignment of the landmarks. The improved registration results are also evident in the much sharper overlaid image produced by our model.}
  \label{fig:geo-registration-results}
\end{figure}

We evaluated RetinaRegNet against leading retinal registration methods using three retinal image registration datasets. In the public FIRE dataset~\citep{hernandez2017fire} of color fundus images, when compared to the best existing method, RetinaRegNet improved the AUC from 0.779 to 0.901 and reduced the mean landmark error from 6.01 to 2.97.
In the public FLoRI21 dataset~\citep{ding2021flori21} of fluorescein angiography images, when compared to the best existing method, RetinaRegNet improved the AUC of the second-best method from 0.640 to 0.868 and reduced the mean landmark error from 36.49 to 13.83.
In the private LSFG dataset of laser-speckle flowgraphy images, when compared to the best existing methods, RetinaRegNet improved the AUC of the second-best method from 0.853 to 0.861 and reduced the mean landmark error from 4.23 to 4.00.
This state-of-the-art performance across various retinal image datasets affirmed RetinaRegNet's significant potential in revolutionizing retinal image registration.
\vspace{5pt}

In summary, this paper makes the following key contributions:
\begin{itemize}
\item We introduced RetinaRegNet, a zero-shot retinal image registration model capable of registering retinal images with large deformations and minimal overlap.
\item We demonstrated RetinaRegNet's state-of-the-art performance across three retinal image registration datasets.
\item We evaluated the significance of each component in RetinaRegNet through comprehensive ablation studies.
\end{itemize}

\section{Related Work}
\subsection{Retinal Image Registration}

Existing retinal image registration methods can be divided into two categories: intensity-based methods and feature-based methods, or a combination of the two~\citep{wang2015robust}. Intensity-based image registration methods typically utilize an intensity similarity function (e.g., mutual information, cross-correlation, sum of squared differences) to align the intensity differences between two images~\citep{zhu2007mutual, legg2013improving, molodij2015enhancing}. While intensity-based registration methods may achieve promising results, they are susceptible to overfitting when dealing with images exhibiting large deformations, minimal overlap, and variations in illumination and texture~\citep{wang2015robust}. Additionally, the choice of the similarity function is task-specific due to variations in imaging modality. In contrast, feature-based registration methods rely on image features such as vasculature bifurcations, the fovea, and the optic disc for transformation estimation. Feature-based methods are generally more robust compared to intensity-based methods. Due to these advantages, numerous feature-based retinal image registration techniques have been developed, using both conventional~\citep{wang2019gaussian, hernandez2020rempe, motta2019vessel} and learning-based approaches~\citep{zou2020non, chen2015retinal, wang2020segmentation, liu2023geometrized, sun2021loftr, lee2019deep, santarossa2022medregnet}. Learning-based techniques outperform conventional methods by achieving higher accuracy, robustness, and efficiency, but they often require extensive preprocessing~\citep{zou2020non, chen2015retinal, wang2020segmentation} or training on large datasets~\citep{liu2023geometrized, sun2021loftr, lee2019deep, santarossa2022medregnet}.

\subsection{Semantic Correspondences}
The goal of semantic correspondence is to identify and match similar semantic features across different images. This process often involves finding correspondences between objects (e.g., points, lines, or regions) based on their semantic meaning rather than visual similarities. Semantic correspondence typically relies on advanced deep learning methods, including convolutional neural networks and vision transformers, to extract and link these semantically similar elements across varied visual representations~\citep{han2017scnet, liu2020semantic, zhao2021multi}. There are two major differences between semantic correspondence and image registration. First, semantic correspondence identifies point correspondences by matching semantic features instead of visual features. Consequently, the images in semantic correspondence can depict two different objects, for example, birds of different species, which are not of interest in image registration. Second, while image registration aims to identify one-to-one point correspondences between two images, semantic correspondence seeks to establish sparse point correspondences (not necessarily one-to-one) between parts of images that share the same underlying features.

\subsection{Diffusion Models}
Denoising diffusion models~\citep{ho2020denoising, song2020score, kingma2021variational, rombach2022high} have emerged as a leading class of generative models renowned for their capacity to generate high-quality images. These models function by iteratively refining an initial Gaussian noise image through a series of reverse diffusion steps, which gradually diminishes noise and enhances image quality. Diffusion models have demonstrated promising performance across various applications, including image denoising~\citep{gong2023pet}, image super-resolution~\citep{li2022srdiff}, image inpainting~\citep{lugmayr2022repaint}, image segmentation~\citep{amit2021segdiff}, and image-to-image generation~\citep{jiang2024fast}.
Training and sampling diffusion models in the original image space can be computationally expensive. A more efficient approach is to conduct the forward and reverse diffusion processes in a smaller latent image space. A prime example of this is stable diffusion~\citep{rombach2022high}, which has subsequently been repurposed for numerous generation tasks such as text-to-3D conversion~\citep{poole2022dreamfusion} and image editing~\citep{mokady2023null}.
Stable diffusion features have recently found applications in representation learning~\citep{yang2023diffusion} and semantic correspondences~\citep{tang2024emergent, hedlin2023unsupervised}.

\section{Method}

\begin{figure*}[!htbp]
  \centering
  \includegraphics[width=1.0\linewidth]{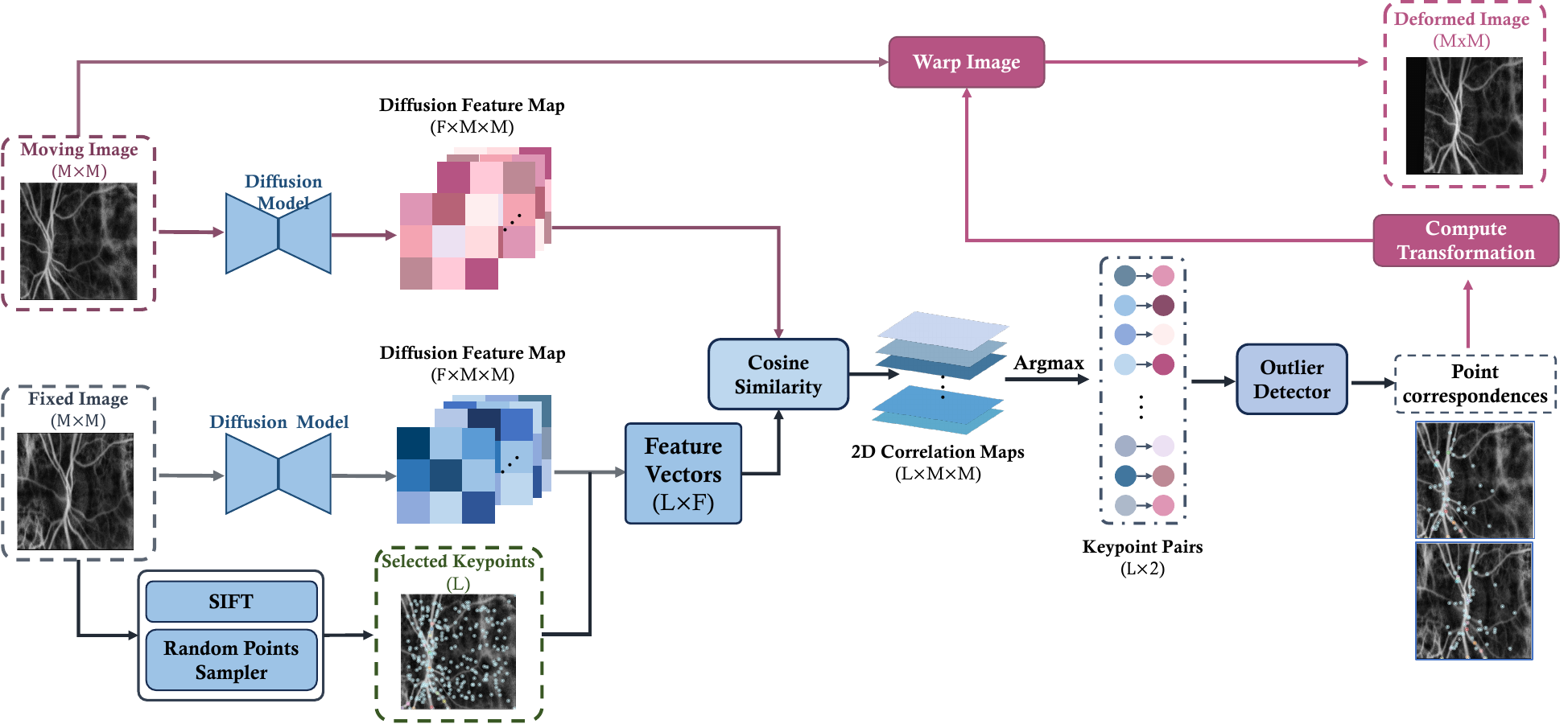}
  \caption{ Overview of the proposed RetinaRegNet model for retinal image registration.}
  \label{fig:RAM-Overview}
\end{figure*}

Figure~\ref{fig:RAM-Overview} provides an overview of our RetinaRegNet image registration model. Inspired by the application of diffusion features (DIFT) in semantic correspondence~\citep{tang2024emergent}, our model began by extracting image features using a pre-trained stable diffusion model~\citep{rombach2022high}. Subsequently, we sampled key feature points in the fixed image using a combination of the SIFT algorithm and random point sampling. This approach allowed for the extraction of feature points in areas with both rich and poor textures. For each sampled feature point in the fixed image, we computed a 2D correlation map to identify its corresponding point in the moving image. To remove outliers in the estimated point correspondences, we used an inverse consistency constraint combined with a transformation-based outlier detector. These strategies were crucial for significantly improving the accuracy and robustness of our registration model.

\subsection{Diffusion Feature Extraction}   
We briefly describe the extraction of diffusion image features proposed in~\citep{tang2024emergent}. The forward diffusion process is a Markov chain that starts with a noise-free image and gradually adds noise to it over $T$ steps, transforming the image into pure Gaussian noise. The model then learns to reverse this process, starting with a pure noise image and gradually removing noise from the image to generate new data samples. In latent diffusion models, the diffusion process primarily occurs in the latent space. At each time step $t$, noise is added to the latent image $z_{t-1}$ to generate $z_t$ with the following equation:
\begin{equation}
q(z_t|z_{t-1}) := \mathcal{N}(z_t; \sqrt{1 - \beta_t} z_{t-1}, \beta_t \mathbf{I})
\end{equation}
where $0<\beta_0<\cdots<\beta_T<<1$ is a noise scheduler.

The reverse process involves training a model $\epsilon_\theta(z_t,t)$, typically implemented as a U-Net, to predict the Gaussian noise $\epsilon$ added at each step of the forward process. The reverse process is represented as:
\begin{equation}
z_{t-1} = \frac{1}{\sqrt{\alpha_t}} \left( z_t - \frac{1 - \alpha_t}{\sqrt{1 - \bar{\alpha}t}} \epsilon_\theta(z_t, t) \right) + \sigma_t \mathbf{z}
\end{equation}
where $\alpha_t = 1 - \beta_t$, $\bar{\alpha}_t = \prod_{s=1}^{t} \alpha_s$, $\sigma^2_t = \frac{1 - \bar{\alpha}_{t-1}}{1 - \bar{\alpha}_t} \beta_t$, and $\mathbf{z} \sim \mathcal{N}(\mathbf{0}, \mathbf{I})$.

To generate diffusion features of a high-quality image $x_0$, we initially encoded the image into the latent representation $z_0$ using the encoder of a pre-trained autoencoder. Then, we introduced random noise to $z_0$ by selecting a time step $t$ and calculating the noisy image using: $z_t = \sqrt{\bar{\alpha}_t} z_0 + (1 - \bar{\alpha}_t) \epsilon_t$, where $\epsilon_t$ represents a random noise sample from $\mathcal{N}(\mathbf{0}, \mathbf{I})$. Subsequently, $z_t$ was fed into the U-Net denoiser, and we extracted features from intermediate layers to serve as diffusion images features.

\subsection{Feature Point Extraction}
The goal of this step was to identify candidate feature points in the fixed image for which we identified corresponding points in the moving image. The SIFT algorithm~\citep{lowe2004distinctive} is a widely used feature extraction algorithm in computer vision for detecting and describing local features in images that are invariant to scaling and rotation. We first employed the SIFT algorithm to detect $K$ feature points in the fixed image, ensuring that the distance between any two feature points was greater than a threshold value, $T_{sift}$. For certain applications, such as retinal images, feature points detected by the SIFT algorithm were concentrated in the sparsely distributed vessel trees, which may have caused the estimated transformation to be less accurate in regions without densely distributed vessels. To address this issue, in addition to the feature points detected by the SIFT algorithm, we randomly sampled an additional $K$ feature points throughout the fixed image to improve the estimation of point correspondences, especially in homogeneous image regions. These points were randomly sampled such that they remained within the image boundary.

\subsection{Estimation of Point Correspondences}
Given a high-resolution fixed image $I_f$ of size $H_f \times W_f$ and a moving image $I_m$ of size $H_m \times W_m$, we first resampled both images to the size of $M\times M$. Diffusion image features were extracted using an intermediate layer of the denoising U-Net in a pre-trained stable diffusion model~\citep{rombach2022high}.
We upsampled the feature maps to the size of $M \times M$. The resulting diffusion features were denoted as $D_f$ and $D_m$, with dimensions $F\times M \times M$, where $F$ represents the size of each feature vector.

We sampled a list of $L = 2K$ feature points $p_1,\cdots, p_{L}$ in the fixed image to find the point in the moving image corresponding to each $p_i$. We computed the cosine similarity score between $D_f(p_i)$ and the feature vector of every pixel in the moving image. The pixel with the highest similarity score was considered to correspond to $p_i$. For computational efficiency, we combined the diffusion feature vectors at all feature points into a matrix $[D_f(p_1), \cdots, D_f(p_L)]$ of size $L\times F$.
We then computed a 2D correlation map for each feature point, resulting in a correlation map $C = [C_1,\cdots,C_L] $ which consisted of $L$ 2D correlation maps:
\begin{equation}
C =  \textbf{norm}([D_f(p_1), \cdots, D_f(p_L)]^T) * \textbf{norm}(D_m)
\end{equation}
where the $\textbf{norm}$ operation was applied to each feature vector to convert it into a vector of unit norm, and $*$ is matrix multiplication.

The correlation map $C$ is of dimension $L\times M \times M$, such that each feature point corresponds to a 2D correlation map. We then applied the Argmax operation to each of the $L$ correlation maps to obtain the corresponding points in the moving image, i.e.,
\begin{equation}
[p'_1,\cdots,p'_L] = [\text{Argmax}(C_1),\cdots, \text{Argmax}(C_L)]
\end{equation}

\subsection{Outlier Detector} 

Point correspondences estimated by the 2D correlation maps can contain errors (outliers) due to variations in image quality. To address this, we designed a two-stage outlier detector that effectively eliminates these outliers from the estimated correspondences, enhancing the robustness of the transformation estimation between image pairs.

\paragraph{\textbf{Inverse Consistency Based Outlier Detector.}}
Inverse consistency~\citep{Pascal1992,christensen2001consistent} was originally used in pairwise image registration to ensure that the forward and backward transformations estimated between two images are inverses of each other. In this paper, we applied this concept to remove outliers in point correspondences between two images. For each feature point $p_i$ in the fixed image, we first determined its corresponding point $p'_i$ in the moving image using a 2D correlation map. Then, we reversed the roles of the fixed and moving images and used the same method to identify the point $p''_i$ in the fixed image that corresponds to $p'_i$. A necessary criterion for the accuracy of estimating point correspondences using 2D correlation maps was the proximity of $p_i$ to $p''_i$. We considered the point correspondence $(p_i, p'_i)$ accurate if it met the following condition; otherwise, we considered this correspondence an outlier and removed it from the estimation of the geometric transformation:
\begin{equation}
||p_i - p''_i||_2 \leq T_{IC},
\end{equation}
where $T_{IC}$ represents the inverse consistency threshold.

\paragraph{\textbf{Transformation Based Outlier Detector.}}
The estimation of a geometric transformation based on point correspondences is sensitive to outliers. The RANSAC algorithm~\citep{fischler1981random} is one of the most widely used outlier detectors. It iteratively selects a random subset of the original data, fits a model, and then evaluates how many of the remaining data points conform to this model within a predefined tolerance. While conceptually straightforward and easy to implement, RANSAC can be computationally intensive for large datasets or complex models. Its performance heavily relies on the chosen threshold for distinguishing inliers from outliers and the number of iterations. 

We observed that the transformation estimation was most sensitive to outliers with large distances between the true corresponding point and the estimated corresponding point. Inspired by this observation, we proposed a simple yet effective transformation-based outlier detector. Given a set of estimated point correspondences ${(p_i, p'_i)}$ between two images, our method first estimates a global transformation (e.g., affine) $\phi$ using all the point correspondences. The global transformation is then applied to each point $p_i$, resulting in transformed points $\phi(p_i)$. The accuracy of each point correspondence is evaluated by computing the Euclidean distance between $\phi(p_i)$ and $p'_i$. Point correspondences with a distance greater than a threshold $T_{trans}$ were removed. This process can be summarized by the following:
\begin{equation}
 (p_i,p'_i) \text{ is an outlier if } || \phi(p_i) - p'_i ||_2 \geq T_{trans}
\end{equation}

\subsection{Multi-Stage Image Registration Framework}
We designed our registration model as a multi-stage framework to handle large variations in the underlying transformation between images through a coarse-to-fine strategy. In the first stage, we estimated a global transformation $\psi_{global}$ between the two images, where estimating such a global transformation is less sensitive to outliers than a more complex local transformation. In the second stage, we estimated a local deformation $\psi_{local}$ between the globally aligned image pairs. The composition of these two transformations gave the final transformation between the two images, i.e., $\psi = \psi_{local}\circ\psi_{global}$.
We have implemented four distinct types of transformation models: affine transformation (6 degrees of freedom), homography transformation (8 degrees of freedom), quadratic transformation (12 degrees of freedom), and third-order polynomial transformation (20 degrees of freedom).
For all experiments in this study, we chose the homography transformation model in the first stage for global alignment, followed by the third-order polynomial transformation model in the second stage to refine local feature alignment.

\vspace{-8 pt}

\section{Experiments}

\subsection{Datasets}
We used three retinal image registration datasets for model evaluation: color fundus, fluorescein angiography, and laser speckle flowgraphy. These different modalities capture retinal images for various applications, including diagnosing retinal conditions, assessing treatment outcomes, and monitoring retinal blood flow dynamics.

\paragraph{\textbf{FIRE Dataset}} The first dataset we used is the public FIRE (Fundus Image Registration Dataset) dataset~\citep{hernandez2017fire} that contains 129 color fundus images that form 134 image pairs. 
Acquired using a Nidek AFC-210 fundus camera at the Papageorgiou Hospital from 39 patients, the dataset offered a resolution of 2912$\times$2912 pixels with a field of view (FOV) of 45°.
The FIRE dataset provided ground truth correspondences for 10 landmark points for each image pair. These landmark points were manually selected by an annotator, primarily focusing on vessels and crossings, to ensure broad coverage of the overlapping image areas. Image pairs in the FIRE dataset were divided into three categories. Class $S$ consists of 71 image pairs characterized by high spatial overlap (greater than 75\%) and no significant visual anatomical differences, making them easy to register. Class $P$ consists of 49 pairs with a small overlap (less than 75\%), making these pairs hard to register. Finally, Class $A$ includes 14 image pairs with considerably large overlap. This  exhibits visual anatomical differences which make them moderately hard to register.

\paragraph{\textbf{FLoRI21 Dataset}} The second dataset we used was the public FLoRI21 (Fluorescein-angiography Longitudinal Retinal Image 2021) dataset~\citep{ding2021flori21} which consists of 15 pairs of ultra-wide-field fluorescein angiography (FA) images taken 24 weeks apart for each subject.
These images were captured using Optos California (Nikon Co. Ltd, Japan) and 200Tx cameras.
Each image pair had one montage FA image (4000 × 4000) used as the fixed image, and several raw FA images (3900 × 3072) used as the moving images. 
 Ten pairs of landmark points were chosen at retinal vessel bifurcations, ensuring coverage of the entire overlapping image area.

\paragraph{\textbf{LSFG Dataset}} The third dataset we used was an Institutional Review Board-approved laser speckle flowgraphy (LSFG) dataset that contains 15 pairs of longitudinal LSFG images from patients with uveal melanoma. These patients were treated at the University of Iowa with $^{125}$I-plaque brachytherapy. LSFG (Softcare Co., Japan) is non-invasive and can measure mean blur rate, which is linearly proportional to blood flow velocity at the retina and optic nerve head in a video clip of 118 time frames. We averaged each pixel value along the aligned time-sequence frames of the LSFG video to create a 2D blood flow map (751 $\times$ 420). Each LSFG image pair has 6-10 ground truth landmark correspondences which were selected by a human annotator.

\subsection{Methods for Comparison}
We compared our RetinaRegNet model with six recent image registration methods. Among these, the GFEMR algorithm~\citep{wang2019gaussian} tackles the registration problem through a probabilistic approach which integrates manifold regularization to preserve the retina's intrinsic geometry. Another method, ASpanFormer~\citep{chen2022aspanformer}, uses an adaptive span transformer to find point correspondences. This employs CNN encoders and iterative global-local attention blocks which utilize auxiliary flow maps to adapt local attention span based on matching uncertainty. Another method we considered was SuperGlue~\citep{sarlin2020superglue}, a neural network that matches two sets of local features by jointly finding correspondences and rejecting non-matchable points. Additionally, SuperRetina~\citep{liu2022semi}, employs a trainable keypoint detector and descriptor. It is trained using semi-supervised learning with a combination of labeled and unlabeled images; however, it requires manual or automatic labeling and is subject to keypoint detection errors and mismatches. Moreover, we also considered the LoFTR algorithm~\citep{sun2021loftr}, which uses a transformer network to create feature descriptors from both images, establishing dense correspondences at a coarse level before refining them. Lastly, GeoFormer~\citep{liu2023geometrized}, another general-purpose method based on LoFTR, leverages RANSAC geometry to identify attentive regions and uses the transformer's cross-attention mechanism for feature matching during transformation estimation.

\subsection{Evaluation Metrics}

\paragraph{\textbf{Mean Landmark Error}}Given a pair of fixed and moving images, we have a set of $N$ manually annotated landmark pairs. Each landmark point in one image is denoted as ($x_i$, $y_i$), and its corresponding ground truth landmark point in the other image is denoted as ($x_i'$, $y_i'$). Let $\psi$ represent a geometric transformation that we have estimated to align the fixed image with the moving image. The mean landmark error (MLE) for this pair of images is defined as the mean of the Euclidean distance between each transformed point $(x''_i, y''_i) := \psi(x_i,y_i)$ and its corresponding ground truth point ($x_i'$, $y_i'$). Mathematically, the MLE can be expressed as:
\begin{equation}
    MLE = \frac{1}{N} \sum_{i= 1}^{N} \sqrt{(x''_i - x'_i)^2 + (y''_i - y'_i)^2}
\end{equation}

\paragraph{\textbf{Area Under the Curve}}The normalized area under the curve (AUC) evaluates an image registration method by assessing its performance at various MLE thresholds ($[0, T_{AUC}]$). For each threshold value, a registration is considered accurate if its MLE is below the threshold. A registration accuracy, indicating the proportion of accurately registered images, is computed for the entire image set at each threshold. The normalized AUC metric integrates these success rates across all MLE thresholds which provides a comprehensive assessment of registration accuracy. Mathematically, the normalized AUC can be expressed as
\begin{equation}
    AUC = \frac{1}{T_{AUC}}\int_{0}^{T_{AUC}} RA(T) dT
\end{equation}
where $RA(T)$ denote the registration accuracy at the threshold value $T$.

\paragraph{\textbf{Registration Success Rate}} We categorize the image registration of an image pair as unsuccessful under either of two conditions. Firstly, the registration is considered failed if there are insufficient point correspondences to estimate the required geometric transformation. For example, a minimum of 3 pairs of points is necessary to estimate an affine transformation, and at least 6 pairs are needed for a quadratic transformation. Secondly, the registration is considered unsuccessful if the mean landmark error exceeds a predetermined threshold, $T_{SR}$, which is determined by the specific characteristics of the registration dataset.
By applying these criteria, we can calculate the success rate of each registration method across different datasets. 
In this paper, we chose $T_{SR} = \frac{1}{2}T_{AUC}$.

\subsection{Implementation Details}

The proposed model was implemented using PyTorch and Diffusers on a computing node equipped with two CPU cores, 25 GB of RAM, and a NVIDIA A100 GPU. For image registration, images from the FIRE, FLoRI21, and LSFG datasets were resampled to resolutions of $920 \times 920$, $1024 \times 1024$, and $740 \times 740$ pixels, respectively. After registration, the coordinates of corresponding points were upsampled to their original image sizes for transformation estimation and landmark error evaluation.We extracted $K = 1000$ feature keypoints from the moving image using the SIFT algorithm, ensuring that the distance between any two keypoints was greater than 10 pixels. Additionally, we randomly selected another set of $K = 1000$ feature points from the moving image.Diffusion image features were extracted using the third layer of the denoising U-Net at time step $t = 1$ for each dataset. The inverse consistency threshold was set at $T_{IC} = 3$ pixels for all datasets. For the computation of the AUC, threshold values ($T_{AUC}$) were set at 25, 100, and 25 for the FIRE, FLoRI21, and LSFG datasets, respectively. For the transformation-based outlier detector, we selected the affine transformation model for both registration stages and set threshold values ($T_{trans}$) to 25 and 15 for the FIRE dataset, 40 and 30 for FLoRI21, and 25 and 25 for the LSFG datasets, respectively. A higher threshold was selected for the FLoRI21 dataset to accommodate its predominantly non-affine deformation in image pairs.

\section{Results}

\subsection{Quantitative Evaluation}

\paragraph{\textbf{Results on the FIRE Dataset.}}

Table~\ref{tab:my_label} demonstrates that our model significantly outperformed all other methods in the challenging $P$ class of hard-to-register cases, delivering the highest AUC of 0.856, compared to the 0.697 AUC of the second-best method, GFEMR. 
These results underscore our model's superior ability to handle large displacement deformations and small overlaps between images. Furthermore, in the other two classes, our model achieved state-of-the-art performance, significantly improving the overall AUC across the FIRE dataset from 0.779 (achieved by SuperRetina) to 0.901, and reducing the mean landmark error from 6.01 to 2.97. Additionally, our model proved to be the most robust, achieving a registration success rate of 99.25\%, significantly surpassing the 91.04\% success rate of GFEMR. 

\paragraph{\textbf{Results on the FLoRI21 Dataset}} 
Challenges in registering image pairs in the FLoRI21 dataset are primarily from the large displacement and scaling deformations between the montage and raw FA images. The results presented in Table \ref{tab:flori21-results} show that our model significantly outperformed the second-best performing model, GeoFormer, in several key metrics. Specifically, our model achieved an AUC of 0.868 compared to GeoFormer's 0.640, a mean landmark error of 13.83 versus 36.49, and a 100\% registration success rate, substantially higher than GeoFormer's 93.33\%. It was notable that AspanFormer, GFEMR, LoFTR, SuperGlue, and SuperRetina all struggled to achieve accurate registration in at least 20\% of the cases. In particular, GFEMR, which is customized for registering color fundus images, was not well-suited for addressing the large scaling and displacement deformations in another retinal dataset. This resulted in a high mean landmark error of 71.58 and a low success rate of 6.67\%. 

\paragraph{\textbf{Results on the LSFG Dataset.}} 
Image pairs in the LSFG dataset represent longitudinal measurements of blood flow in the eyes. The primary challenges in registering these LSFG image pairs arise from two factors: (1) positional shifts across images, attributable to the different times at which they were acquired, and (2) intensity variations across images, due to changes in blood flow over time. The results, as detailed in Table \ref{tab:lsfg-results}, demonstrate that our method outperformed all other methods in several key metrics. Specifically, our model achieved an AUC of 0.861, compared to the second-best method, LoFTR's 0.853, and a mean landmark error of 4.00, slightly better than LoFTR's 4.23. 

\paragraph{\textbf{Computational Complexity}} 
For each retinal registration dataset, we computed the average per-case running time of each image registration method. Results in Table~\ref{tab
} suggest that, although RetinaRegNet achieved state-of-the-art performance on all three retinal registration datasets, it is the slowest method, taking up to 20 seconds to register each image pair. This extended running time is primarily due to the high computational cost associated with extracting diffusion features and computing 2D correlation maps for each feature point.

\begin{table}[!hbt]
    \centering
    \caption{Registration results on the FIRE Dataset. MLE: mean landmark error. Success rate threshold: \( T_{SR} = 12.5 \).}
    \label{tab:my_label}
    \scalebox{0.75}{
        \begin{tabular}{ccccccc}
            \toprule
            \textbf{Method} & Easy & Moderate & Hard & FIRE & MLE & Success Rate \\
            \midrule
            GFEMR & 0.849 & 0.534 & 0.697 & 0.761 & 6.11 & 91.04\% \\
            ASpanFormer & 0.878 & 0.728 & 0.057 & 0.562 & 18.05 & 64.92\% \\
            SuperGlue    & 0.810 & 0.602& 0.417 & 0.645 & 9.59 &70.89\% \\
            LoFTR & 0.941 & 0.746 & 0.338 & 0.703 & 8.29 & 74.06\% \\
            SuperRetina & 0.946 & 0.771 & 0.539 & 0.779 & 6.01 & 83.58\% \\
            GeoFormer & 0.922 & 0.757 & 0.564 & 0.774 & 6.14 & 88.80\% \\
            \midrule
            \textbf{Ours} & \textbf{0.951} & \textbf{0.805} & \textbf{0.856} & \textbf{0.901} & \textbf{2.97} & \textbf{99.25\%} \\
            \bottomrule
        \end{tabular}
    }
\end{table}

\begin{table}[!htb]
    \centering
    \caption{Registration results on the FLoRI21 Dataset. MLE: mean landmark error. Success rate threshold: \( T_{SR} = 50 \).}
    \label{tab:flori21-results}
    \scalebox{0.75}{
        \begin{tabular}{c c c c}
            \toprule[0.75 pt]
            \textbf{Method} &  AUC    &   MLE    &   Success Rate \\
            \midrule[0.5pt] 
            GFEMR        & 0.299 & 71.58 &  6.67\% \\ 
            ASpanFormer  & 0.548 & 45.59 & 60.00\%\\
            SuperGlue     & 0.604 & 40.09& 80.00\% \\
            LoFTR& 0.486 &  51.99 & 60.00\%\\ 
            SuperRetina         & 0.590 & 41.47 & 80.00\%  \\ %
            GeoFormer     & 0.640 & 36.49 & 93.33\%  \\ %
            \midrule[0.1pt] 
            \textbf{Ours}                         & \textbf{0.868} & \textbf{13.83} & \textbf{100\%} \\ 
            \bottomrule[0.95 pt]
        \end{tabular}
    }
\end{table}

\begin{table}[!htb]
    \centering
    \caption{Registration results on the LSFG Dataset. MLE: mean landmark error. Success rate threshold: \( T_{SR} = 12.5 \).}
    \label{tab:lsfg-results}
    \scalebox{0.75}{
        \begin{tabular}{c c c c}
            \toprule[0.95 pt]
            \textbf{Method} &  AUC    &   MLE    &   Success Rate   \\
            \midrule[0.5pt]
            GFEMR      &  0.587   & 15.29  & 73.33\%  \\
            ASpanFormer  & 0.813 & 5.07 & 93.33\%\\
            SuperGlue     & 0.752 & 6.66 & 93.33\% \\
            LoFTR  & 0.853 & 4.23 & 100\%\\
            SuperRetina         & 0.843 & 4.32 & 100\%  \\
            GeoFormer     & 0.845 & 4.40 & 100\% \\
            \midrule[0.1pt] 
            \textbf{Ours}            &  \textbf{0.861}          &  \textbf{4.00}  & \textbf{100\%} \\
            \bottomrule[0.95 pt]
        \end{tabular}
    }
\end{table}

\begin{table}[!htbp]
    \centering
    \caption{Per-case computational complexity (seconds).}
    \label{tab:complexity-results}
    \scalebox{0.75}{
        \begin{tabular}{c c c c}
            \toprule[0.95 pt]
            \textbf{Method} & FIRE &  FLoRI21    &  LSFG     \\
            \midrule[0.5pt] 
            GFEMR  &  3.032  & 5.313  & 0.9211  \\
            ASpanFormer  & 1.00 & 1.00 &  1.00\\
            SuperGlue     & 1.00 & 4.00 & 5.00 \\
            LoFTR  &  0.363 &  \textbf{0.508} &  0.324\\
            SuperRetina   & 1.172 & 2.157 & 1.148 \\
            GeoFormer  & \textbf{0.36} & 0.53 &  \textbf{0.18} \\
            \midrule[0.1pt] 
            \textbf{Ours}            &  13.00          &  19.00  & 18.00 \\
            \bottomrule[0.95 pt]
        \end{tabular}
    }
\end{table}

\subsection{Qualitative Evaluation}
\begin{figure}[h!]
  \centering
  \includegraphics[width=1.0\linewidth]{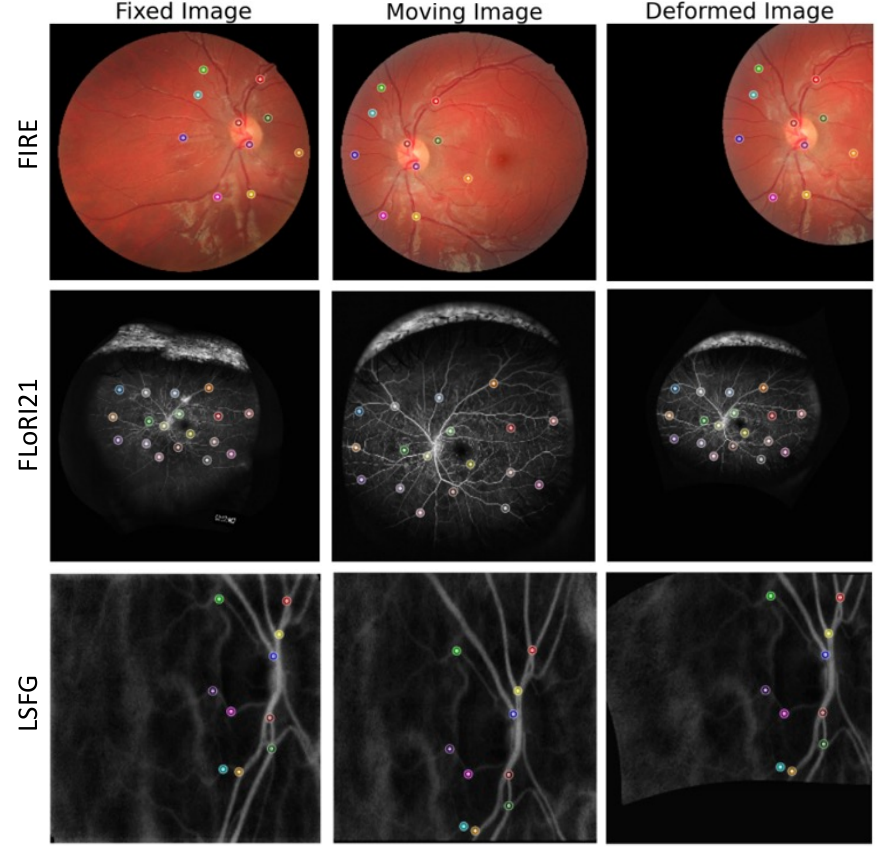}
  \caption{Registration results of the model are shown for the FIRE (first row), FLoRI21 (second row), and LSFG (third row) datasets. For each landmark, we assigned a unique color, consistent across fixed, moving, and deformed images, enabling clear observation of the alignment of each landmark pair.}
  \label{fig:registration-results}
\end{figure}

Figure~\ref{fig:registration-results} shows the registration results of one representative case from each of the three retinal image registration datasets. The corresponding landmarks in the fixed, moving, and deformed images demonstrate the high accuracy of our registration model across all three cases. The first row demonstrates that our model has successfully estimated the large displacement deformation between an image pair from the hard-to-register ($P$) category in the FIRE dataset. The second row shows how our model has accurately aligned an image pair with complex global and local deformations from the FLoRI21 dataset. The last row demonstrates the superior performance of our model on a pair of images in our LSFG dataset with large shifting. The consistently high performance of our model across all three datasets demonstrates its accuracy, robustness, as well as its potential to be utilized in other retinal image registration tasks.

\section{Ablation Study}

\subsection{Impact of Image Feature Extractor}

We conducted an in-depth evaluation of the effectiveness of using different feature types in RetinaRegNet, including CNN features~\citep{simonyan2014very}, vision transformer features~\citep{caron2021emerging}, and diffusion features~\citep{tang2024emergent}. Specifically, we utilized a pretrained VGG19 model for extracting CNN features, a pretrained 8-patch vision transformer network (ViT-S/8) for extracting DINO-ViT features, and a pretrained latent stable diffusion model (SDv2-1) for extracting diffusion features. We applied RetinaRegNet to re-register all pairs of images by replacing the diffusion features with the CNN features and DINO-ViT features. Registration results in Table~\ref{tab
} suggested that diffusion features achieved the best performance across all three datasets. RetinaRegNet achieved similar registration results on the FIRE and FLoRI21 datasets when using CNN features and diffusion features. However, the AUC of RetinaRegNet decreased from 0.861 to 0.791 on the LSFG dataset when using CNN features instead of diffusion features. Notably, the DINO-ViT features did not exhibit the same level of accuracy compared to the CNN features and diffusion features. Therefore, we chose to use diffusion features (DIFT) in the proposed RetinaRegNet registration framework.

\begin{table}[!htbp]
    \centering
    \caption{AUC values of RetinaRegNet using different image feature extractors.}
    \label{tab:ablation_feature_extraction}
    \scalebox{0.75}{
        \begin{tabular}{c c c c}
            \toprule[0.95 pt]
            \textbf{Features} & FIRE &  FLoRI21    &  LSFG     \\
            \midrule[0.5pt] 
             CNN Features & {0.894}  & {0.860} & {0.791} \\
             DINO-ViT Features & {0.410} & {0.537} & {0.199} \\
             Diffusion Features & \textbf{0.901} & \textbf{0.868} & \textbf{0.861}\\
            \bottomrule[0.95 pt]
        \end{tabular}
    }
\end{table}

\subsection{Impact of Diffusion Feature Size}
Image features from various layers of the decoder in the stable diffusion model exhibit differing sizes, each smaller than the input image by a downsampling factor. We assessed how variations in diffusion feature size affect RetinaRegNet's performance. Ideally, image features should capture high-level features without being either too shallow or too deep. Shallowness corresponds to a small downsampling factor with limited learned representations and deepness, corresponds to a large downsampling factor unsuitable for dense predictions due to excessively small image features.

Results in Table \ref{tab:ablation_diffusion_features} indicated that the diffusion features corresponding to a downsampling factor of 32 exhibited the best registration performance across all three retinal datasets; these features provide a balance between detailed representation and abstraction. Diffusion features corresponding to a downsampling factor of 16 also achieved promising registration results, ranking second best. Notably, the features extracted from the first block and the last block both resulted in poor performance. These blocks represent extremely low and high-level diffusion features, respectively.

\begin{table}[!htbp]
    \centering
    \caption{AUC values of RetinaRegNet using diffusion features extracted by different downsampling factors.}
    \label{tab:ablation_diffusion_features}
    \scalebox{0.75}{
        \begin{tabular}{c c c c c c}
            \toprule[0.95 pt]
            \textbf{Feature Selection} & Downsampling Factor & FIRE &  FLoRI21    &  LSFG     \\
            \midrule[0.5pt] 
             Block-1 Features & {8}  & {0.086} & 0.000 & 0.000 \\
             Block-2 Features & {16}  & {0.742} & {0.796} & {0.834} \\
             Block-3 Features & \textbf{32}  & \textbf{0.901} & \textbf{0.868} & \textbf{0.861}\\
             Block-4 Features & {64}  & {0.554} & {0.301} & {0.051} \\
            \bottomrule[0.95 pt]
        \end{tabular}
    }
\end{table}

\subsection{Impact of Key Model Components}
We investigated the impact of each of the four key components in RetinaRegNet, including the inverse consistency constraint, the random feature point sampling strategy, the transformation-based outlier detector, and the multi-stage registration framework. We reran RetinaRegNet on all image pairs using one of the following: (1) removing the inverse consistency constraint; (2) omitting the random sampling strategy; (3) replacing the affine transformation-based outlier detector with the RANSAC outlier detector; and (4) replacing the multi-stage framework with a single-stage design.

Results in Table \ref{tab:ablation} indicate that all four components significantly contribute to the high performance of the RetinaRegNet model, although their impact varies across different retinal datasets. For instance, in both the FIRE dataset and the FLoRI21 dataset, the multi-stage image registration design notably improved performance by a significant margin, likely due to the complexity of deformations between image pairs in those two datasets. Conversely, in the LSFG dataset, substituting our affine transformation-based outlier detector with the RANSAC detector resulted in a significantly decreased AUC from 0.861 to 0.416, possibly due to the simpler nature of deformations in these images.

\begin{table}[!htbp]
    \centering
    \caption{Impact of each model component on AUC values.}
    \label{tab:ablation}
    \scalebox{0.75}{
        \begin{tabular}{c c c c}
            \toprule[0.95 pt]
            \textbf{Method} & FIRE &  FLoRI21    &  LSFG     \\
            \midrule[0.5pt] 
            RetinaRegNet (proposed)      &   \textbf{0.901}  & \textbf{0.868} & \textbf{0.861} \\
            RetinaRegNet w/o  inverse consistency & 0.884 &0.847 & 0.850\\
            RetinaRegNet w/o random point sampling & 0.890 & 0.866 & 0.842 \\
            RetinaRegNet w/o affine outlier detector    &   0.805  & 0.835 & 0.416  \\
            RetinaRegNet w/o multi-stage design     &  0.782   & 0.649 & 0.853 \\
            \bottomrule[0.95 pt]
        \end{tabular}
    }
\end{table}

\section{Discussion}

\subsection{Clinical Implications}
The retina, a light-sensitive layer at the back of the eye, plays a vital role in our vision~\citep{kawamura2012explaining}. However, retinal diseases like diabetic retinopathy, glaucoma, and age-related macular degeneration~\citep{duh2017diabetic, imran2022unified, chan2017retinal, lim2012age} can lead to vision impairment. Early detection and treatment are crucial, emphasizing the need for monitoring retinal blood vessels. Retinal image registration, a technique used to align retinal images, facilitates accurate diagnosis and disease monitoring. By precisely tracking changes over time and assessing treatment response, clinicians can effectively manage retinal pathologies. Our RetinaRegNet, achieving state-of-the-art registration accuracy, holds immense potential to revolutionize clinical practice by empowering clinicians with enhanced diagnostic tools and treatment monitoring capabilities, ultimately leading to improved patient outcomes.

\subsection{Limitations and Future Directions}
Our study has several limitations. First, RetinaRegNet is computationally expensive, requiring between 10 to 20 seconds per image pair due to the computation of a 2D correlation map for each selected feature point in the moving images. In this paper, we opted for 2000 feature points (1000 for SIFT; 1000 for random point sampling), however users have the flexibility to choose a lower number of points, particularly when the underlying transformation between the two images is an affine or homography transformation. We will assess the trade-off between computational efficiency and registration accuracy in our future work. Second, our focus was solely on mono-modal image registration, which overlooked the generalization of our method to multi-modal retinal image registration challenges. We plan to evaluate the effectiveness of RetinaRegNet in multi-modal registration by initially training and employing an image-to-image translation network to unify different modalities into a single modality, followed by applying RetinaRegNet. Third, theoretically, RetinaRegNet can serve as a general image registration approach that can be extended to various other registration tasks. In the future, we aim to evaluate RetinaRegNet's performance in other medical image registration contexts (e.g., registering 2D histopathology and MRI images) and nature image registration contexts (e.g., aligning images captured by unmanned aerial vehicles from different angles or times).

\section{Conclusion}

We introduced RetinaRegNet, a zero-shot image registration model designed to address the challenges of large deformations, minimal overlap, and varying image quality in retinal images. Through a novel combination of feature extraction using diffusion models, inverse consistency, outlier detection, and a two-stage registration framework, RetinaRegNet consistently outperformed state-of-the-art methods across three retinal image registration datasets. Our comprehensive evaluation demonstrated its robust performance in registering images across different modalities, highlighting its potential for clinical applications such as disease monitoring and treatment evaluation. Despite some computational limitations, RetinaRegNet represents a significant advancement in retinal image registration.

\printcredits

\section*{Conflict of interest statement}
The authors have no conflict of interest to declare.

\section*{Declaration of generative AI and AI-assisted technologies in the writing process}
During the preparation of this work the authors used the ChatGPT-3.5 and ChatGPT-4.0 tools in order to improve the readability and language of our paper. After using this tool/service, the authors reviewed and edited the content as needed and take full responsibility for the content of the publication.

\section*{Acknowledgments}
This work was supported by the Department of Medicine and the Intelligent Clinical Care Center at the University of Florida College of Medicine.
We would like to express our gratitude to Alexander Cosman for editing the language of this paper.

\bibliographystyle{apalike}


\bibliography{biblioentropy} 

\end{document}